\documentclass[numbers,preprint,12pt]{elsarticle}

\usepackage{natbib}
\usepackage{caption}
\usepackage{booktabs}
\usepackage{array}
\usepackage{subcaption}
\usepackage[ruled]{algorithm2e}
\usepackage{tabularx}
\usepackage{threeparttable}
\usepackage{multirow}
\usepackage{makecell}
\usepackage{etoolbox}
\usepackage{url}
\usepackage{color}

\usepackage{amssymb}
\usepackage{amsmath}
\usepackage{amsthm}

\graphicspath{ {./images/} }

\newcolumntype{C}{>{\centering\arraybackslash}X}
\newcolumntype{P}[1]{>{\centering\arraybackslash}p{#1}}

\begin{document}

\begin{frontmatter}



    \title{Temporal Regularization Training: Unleashing the Potential of Spiking Neural Networks} 


    \author[1]{Boxuan Zhang}

    \author[2]{Zhen Xu}

    \author[1]{Kuan Tao\corref{cor1}}
    \ead{taokuan@bsu.edu.cn}

    \cortext[cor1]{Corresponding author}

    \affiliation[1]{organization={School of Sports Engineering, Beijing Sport University},
        city={Beijing},
        postcode={100091},
        country={China}}

    \affiliation[2]{organization={School of Mathematics, Statistics and Mechanics, Beijing University of Technology},
        city={Beijing},
        postcode={100124},
        country={China}}

    \begin{abstract}
        Spiking Neural Networks (SNNs) have received widespread attention due to their event-driven and low-power characteristics, making them particularly effective for processing neuromorphic data. Recent studies have shown that directly trained SNNs suffer from severe temporal gradient vanishing and overfitting issues, which fundamentally constrain their performance and generalizability. This paper unveils a temporal regularization training (TRT) memthod, designed to unleash the generalization and performance potential of SNNs through a time-decaying regularization mechanism that prioritizes early timesteps with stronger constraints. We perform theoretical analysis to reveal TRT's ability on mitigating the temporal gradient vanishment. To validate the effectiveness of TRT, we conduct experiments on both static image datasets and dynamic neuromorphic datasets, perform analysis of their results, demonstrating that TRT can effectively mitigate overfitting and help SNNs converge into flatter local minima with better generalizability. Furthermore, we establish a theoretical interpretation of TRT's temporal regularization mechanism by analyzing the temporal information dynamics inside SNNs. We track the Fisher information of SNNs during training process, showing that Fisher information progressively concentrates in early timesteps. The time-decaying regularization mechanism implemented in TRT effectively guides the network to learn robust features in early timesteps with rich information, thereby leading to significant improvements in model generalization.
    \end{abstract}




\end{frontmatter}

\section{Introduction}
Spiking neural networks (SNNs) \citep{maass1997snn} are considered as the third generation of neural network models. The biology-inspired binary spike transmitted mechanism brings SNNs multiple advantages and high potential for fast inference and low computation cost on neuromorphic architectures. On the other hand, the non-differentiable spike generation method embodied in this mechanism hinders the widely used back-propagation approaches, making SNNs difficult to train. In recent studies, the artificial neural networks (ANNs) to SNNs conversion method \citep{deng2021optimal,li2021free,lv2024optimal} has been proved an effective approach to obtain high-performance SNNs, while it suffers from the long inference latency \citep{han2020rmpsnn}. Another training approach is to utilize surrogate gradients \citep{lee2016training,wu2018spatio,neftci2019surrogate} and back-propagation through time \citep{neftci2019surrogate} to train high-performance SNNs, which is widely known as direct training of SNNs. Direct training has shown superior computational efficiency, making it the mainstream for training SNNs and propelling its widespread application in various scenarios such as robotics \citep{abadia2019robot}, edge devices \citep{sharifshazileh2021electronic}, brain-machine interfaces \citep{liu2024proliferation,zhu2025adaptive}, etc. In particular, SNNs have shown promising results in processing neuromorphic event-based data due to their ability on processing temporal information \citep{xing2020new,chen2020event}.\par
However, previous studies have shown that SNNs suffer from severe overfitting issues on neuromorphic datasets. Neuromorphic datasets are collected by dynamic vision sensor (DVS) cameras. Unlike traditional frame-based datasets, neuromorphic datasets record asynchronous, sparse temporal events triggered by brightness changes \citep{li2022aug}. Due to the high cost of data collection \citep{lin2021esimagenet}, existing neuromorphic datasets are generally limited in scale, which often leads to overfitting and limits the performance of SNNs \citep{he2024transferdvs}. Although approaches such as data augmentation \citep{li2022aug,qiu2024gated,gu2024eventaugment}, dropout \citep{srivastava2014dropout}, and weight decay \citep{krogh1991simple} can help alleviate this problem, their efficacy is still limited. Moreover, the adoption of a surrogate function leads to gradient mismatching \citep{wang2023adaptive}, which prevents the network from learning with the precise gradients of the loss landscape in SNNs, limiting their performance and making them easily get trapped into the local minimum with low generalization, thereby further exacerbates overfitting issues. To address this issue, \citet{deng2022tet} proposed the temporal efficient training (TET) method, which effectively prevents the model from converging to the local minima with low generalizability, while the ability of TET to mitigate overfitting is still limited and needs further improvement. Some previous work also linked weight decay regularization to the temporal dynamics of SNNs and proposed enhanced temporal regularization methods \citep{panda2017asp,putra2021spikedyn,araki2023supervised,sakemi2023sparse}.
However, the mechanisms underlying their effectiveness are yet to be systematically explained. Additionally, direct training method suffers from the vanishing of temporal gradients on early timesteps \citep{meng2023bptt}, which also limits the performance potential of SNNs.\par
In this work, we first analyze the limitation of the standard direct training (SDT) approach with surrogate gradient, and introduce a novel temporal regularization training (TRT) algorithm to unleash the latent generalization capacity. By integrating a time-dependent regularization mechanism, TRT implements constraints for model's output on every moment, effectively mitigating overfitting and enhancing model generalization. Our analysis reveals that TRT can compensate the vanishing of temporal gradients during training, thereby boosting the SNN performance. Experiments on multiple datasets demonstrate the effectiveness of TRT. Furthermore, we analyze the temporal information dynamics inside SNNs during the TRT training process. We provide an interpretation of the time-dependent regularization in TRT from the perspective of temporal information concentration (TIC) phenomenon and demonstrate that this interpretation generalizes to all time-decaying regularization methods.\par
The main contributions of this work can be summarized as follows:
\begin{itemize}
    \item We propose the TRT method, a new loss function for direct training that provides a time-dependent regularization mechanism to enforce stronger constraints on early timesteps. Theoretical analysis reveals TRT's capability to mitigate temporal gradient vanishing, thereby enhance model performance.
    \item Experiments demonstrate the proposed TRT method achieves state-of-the-art performance on both static datasets and neuromorphic datasets, which proves the effectiveness of TRT.
    \item We picture the loss landscape and the learning curves of TRT to demonstrate its advantage in mitigating overfitting and better generalization.
    \item We visualize the Fisher information across the TRT training epochs. By analyzing the temporal dynamic changes of Fisher information during the training process, we further establish a theoretical interpretation of TRT.
\end{itemize}

\section{Related Work}
\paragraph{\textbf{Conversion methods}} In order to fully leverage the knowledge acquaired from the pre-trained ANNs, conversion methods directly transfer the parameters from the ANN model to the SNN model with the same architecture \citep{diehl2015fast,rueckauer2017ann2snn}. Conversion methods obtain high-performance SNNs by training corresponding ANNs, avoiding the differentiation challenges inherent to the spike generation function. Previous studies have proposed numerous effective improvements to conversion methods, achieving nearly lossless accuracy conversion \citep{diehl2015fast,deng2021optimal}. However, conversion methods often involve long latency and fail to retain the spatiotemporal characteristics inherent to spike-based processing, significantly limiting the practical applicability of converted SNNs \citep{yu2025temporal}.
\paragraph{\textbf{Direct training methods}} Direct training essentially treats SNNs as RNNs and trains them with surrogate gradients and BPTT \citep{neftci2019surrogate}. During the forward phase, the binary spikes emitted by spiking neurons are generated by the Heavside step function. In the backward propagation phase, the non-differentiable Heavside step function is generally replaced by surrogate gradients \citep{lee2016training}. This foundational framework inspired subsequent advancements, \citet{wu2018spatio} propose the spatial-temporal backpropagation (STBP) method to enable SNN training on ANN programming platform. In order to further improve the efficiency of direct training, innovative approaches \citep{wu2019neunrom,fang2021plif,zheng2021tdbn,shen2024implts} and model architectures \citep{fang2021sew} are proposed to accelerate the convergence of SNNs. However, the adoption of a surrogate function causes SNNs to suffer from the gradient mismatching issue, leading to low generalization. Therefore, many techniques \citep{deng2022tet,wang2023adaptive,shen2024implts} have been developed to enhance model generalization and achieve better performance on static and neuromorphic tasks. In this paper, we provide a new loss function with a special designed time-based regular term for direct training, which can significantly improve the generalization.
\paragraph{\textbf{Regularization methods for SNNs}}
Regularization can effectively alleviate overfitting and improve model generalization. SNN typically employs regularization strategies that commonly used in ANN, such as dropout and weight decay \citep{li2021differentiable,wang2022ltmd}. However, such strategies fail to leverage the rich spatial-temporal dynamics inherent in SNNs, consequently leaving SNNs vulnerable to severe overfitting issues. Therefore, conventional regularization strategies are not applicable and require special design \citep{panda2017asp,mukhoty2023localzo,zuo2024temporal,yan2022sparsereg}. These advanced regularization strategies help improve the generalizability of SNNs, while their underlying mechanisms remain insufficiently understood and warrant further investigation. Our temporal regularization method focuses on the temporal information dynamics inherent in SNNs and then provide a time-decaying regularization method to constrain the model output on each timestep. In addition, we provide a theoretical interpretation for time-decaying regularization methods.

\section{Preliminary}
\subsection{Leaky Integrate-and-Fire (LIF) Model}
We adopt the iterative form of the Leaky Integrate-and-Fire (LIF) neuron model \citep{wu2018spatio,yin2020effective}. Mathematically, the update process of the membrane potential is
\begin{equation}
    u(t)={\gamma}u(t-1)+X(t),
    \label{eq:lif_charge}
\end{equation}
where $\gamma=\frac{1}{\tau}$ is the leaky factor that drives the membrane potential to decay over time, and $\tau$ is the membrane time constant. $u(t)$ denotes the membrane potential at time t, and $X(t)$ denotes the pre-synaptic input, which is the product of the synaptic weight $\mathbf{W}$ and the spiking input $s(t)$. The neuron will emit a spike whenever the membrane potential suppresses a given threshold $u_{th}$. Every time the neuron fires a spike, $u(t)$ will be reset to a specific value $u_{reset}$, which is called the hard reset. So the firing and reset mechanism can be described as
\begin{equation}
    s(t)={\Theta}\Big(u(t)-u_{th}\Big)
\end{equation}
\begin{equation}
    u(t)=\Big(1-s(t)\Big){\odot}u(t)+s(t){\cdot}u_{reset},
\end{equation}
where $\odot$ denotes the element-wise production operation, and $\Theta$ denotes the Heavside step function:
\begin{equation}
    {\Theta}(x)=
    \left\{
    \begin{array}{ll}
        1, \quad x>0 \\
        0, \quad otherwise.
    \end{array}
    \right.
\end{equation}
The output spike $s(t)$ serves as the post-synaptic output and is subsequently propagated to the next layer of the network.

\subsection{Training Method}
\paragraph{\textbf{Training framework}} We utilize the spatial-temporal backpropagation algorithm \citep{wu2018spatio} to train SNNs:
\begin{equation}
    \frac{\partial \mathcal{L}}{\partial \mathbf{W}}={\sum_{t=1}^T}\frac{\partial \mathcal{L}}{\partial s(t)}\frac{\partial s(t)}{\partial u(t)}\frac{\partial u(t)}{\partial X(t)}\frac{\partial X(t)}{\partial \mathbf{W}},
    \label{eq:stbp}
\end{equation}
where $T$ is the length of simulate time window, and the term $\frac{\partial s(t)}{\partial u(t)}$ is the gradient of non-differentiable Heaviside step function which is typically replaced by surrogate gradients with derivable curves \citep{neftci2019surrogate} of different shapes like triangle \citep{esser2016convolutional} and rectangle \citep{zhang2025dalif} curve, or in other forms such as adaptive surrogate gradient \citep{wang2023adaptive}. In this work, we use a triangle-shaped surrogate function with adjustable width and amplitude \citep{deng2022tet}, which can be defined as
\begin{equation}
    \frac{\partial s(t)}{\partial u(t)}=\frac{1}{\alpha^2}\max(0,\alpha-|u(t)-u_{th}|),
\end{equation}
where $\alpha$ is a hyperparameter that controls the activation range and magnitude of the surrogate gradient, corresponding to the width and amplitude of the triangle-shaped function.
\paragraph{\textbf{Batch normalization}} The Batch Normalization (BN) \citep{ioffe2015bn} method is proposed to accelerate and enhance training stability, since it can smooth the loss landscape \citep{santurkar2018bnhelp} and prevent both gradient explosion and gradient vanishing problem during training. \citet{zheng2021tdbn} proposed threshold-dependent Batch Normalization (tdBN) for SNNs. tdBN can normalize the pre-synaptic input $X(t)$ in both temporal and spatial dimensions.  Due to its high performance, \citet{deng2022tet} and \citet{mukhoty2023localzo} adopted tdBN with extension of the time dimension to the batch dimension and obtained excellent results, which is also adopted in this work. For BN layers, we flatten both the batch and time dimensions into a unified dimension, enabling joint normalization across these two domains.%

\subsection{Temporal Information Dynamics in SNN}
\paragraph{\textbf{Fisher information analysis}}
The Fisher Information Matrix (FIM) can quantify parametric sensitivity through the second-order variation of the KL divergence under parameter perturbations \citep{fisher1925fim}. Given a certain input data, the magnitude of parameter perturbations indicates the amount of information with respect to the corresponding data contained in the model. To analyze the information dynamics of SNN in the time dimension, \citet{kim2023tic} proposed a method to measure the accumulated FIM in SNN over certain time intervals. Given a network’s approximate posterior distribution $\frac{\partial f_\mathbf{W}(O|x_{i\leq t})}{\partial \mathbf{W}}$ with synaptic weight parameters $\mathbf{W}$ and input data $x$ sampled from data distribution $D$, the amount of FIM accumulated from timestep $1$ to timestep $t$ can be calculated as%
\begin{equation}
    \begin{split}
        M_t=\mathbb{E}_{x\sim D,O\sim f_\mathbf{W}(O|x_{i\leq t})}[\nabla_\mathbf{W}\log f_\mathbf{W}(O|x_{i\leq t})\nabla_\mathbf{W}\log f_\mathbf{W}(O|x_{i\leq t})^\mathsf{T}],
    \end{split}
\end{equation}
where $O$ denotes the network output, $x$ denotes the input, $i\in \{1,...,T\}$ represents the index of timestep, and $\mathbf{W}$ denotes the weight parameters.\par
To avoid computational difficulties caused by excessively large FIM, the trace of $M_t$ is widely used as a measurement of accumulated information stored in weight parameters \citep{kim2023tic,kirkpatrick2017overcoming,kleinman2024fim}. Given $N$ training samples, the trace of $M_t$ can be calculated as
\begin{equation}
    I_t=\frac{1}{N}\sum_{n=1}^N\left\|\frac{\partial f_\mathbf{W}(O|x_{i\leq t})}{\partial \mathbf{W}}\right\|^2.
\end{equation}
In the time dimension, the centroid of Fisher information can be measured with a matrix called \textit{Information Centroid (IC)} \citep{kim2023tic}, which can be calculated as
\begin{equation}
    IC=\frac{\sum_{t=1}^TtI_t}{\sum_{t=1}^TI_t}.
\end{equation}

\paragraph{\textbf{Temporal information concentration in SNNs}}
Temporal Information Concentration (TIC) is a phenomenon observed in the training process of SNN \citep{kim2023tic}. Throughout the training progression, the Fisher information exhibits a temporal migration pattern in which its concentration gradually transitions from later to earlier timesteps. This temporal redistribution phenomenon reflects the growing significance of initial phases in the learning process, as critical information becomes increasingly concentrated during the early training stages.

\section{Methodology}
\subsection{Standard Direct Training for SNN with Surrogate Gradients}
We use $O(t)$ to represent the SNN output at time $t$. The standard cross-entropy loss $\mathcal{L}_\mathrm{SCE}$ can be calculated as
\begin{equation}
    \begin{split}
        \mathcal{L}_\mathrm{SCE} & = \mathcal{L}_\mathrm{CE}\Big(\frac{1}{T}\sum_{t=1}^{T}{O(t)},y\Big),
        \\ &= -\sum_{i=1}^{n}\hat{y}_{i}log\left(\mathrm{Softmax}\Big(\frac{1}{T}\sum_{t=1}^{T}O_i(t)\Big)\right),
    \end{split}
    \label{eq:sce}
\end{equation}
where $T$ is the length of simulate time window, $\mathcal{L}_\mathrm{CE}$ denotes the cross-entropy loss, $y$ denotes the real label, $\hat{y}$ denotes the one-hot encoded label, $O_i(t)$ and $\hat{y}_i$ represent the $i^{th}$ component of $O(t)$ and $\hat{y}$, respectively.\par
The gradient of synaptic weight $\mathbf{W}$ can be calculated as
\begin{equation}
    \frac{\partial \mathcal{L}_\mathrm{SCE}}{\partial \mathbf{W}}=\frac{1}{T}\sum_{t=1}^{T}\sum_{i=1}^{n}\Big(\mathrm{Softmax}(\overline{O}_i)-\hat{y}_i\Big)\frac{\partial O_i(t)}{\partial \mathbf{W}},
    \label{eq:sce_W}
\end{equation}
where $\overline{O}_{i}=\frac{1}{T}\sum_{t=1}^{T}O_i(t)$ denotes the average value of $O(t)$ within the total simulation window $T$.\par
Following the previous setting, the standard mean squared error loss $\mathcal{L}_\mathrm{SMSE}$ can be calculated as
\begin{equation}
    \begin{split}
        \mathcal{L}_\mathrm{SMSE} & = \mathcal{L}_\mathrm{MSE}\Big(\frac{1}{T}\sum_{t=1}^{T}{O(t)},\hat{y}\Big),
        \\&= \frac{1}{n}\sum_{i=1}^{n}\left(\Big(\frac{1}{T}\sum_{t=1}^{T}O_i(t)\Big)-\hat{y}_i\right)^2.
    \end{split}
\end{equation}
The gradient of synaptic weight $\mathbf{W}$ can be calculated as
\begin{equation}
    \frac{\partial \mathcal{L}_\mathrm{SMSE}}{\partial \mathbf{W}}=\frac{2}{n}\cdot\frac{1}{T}\sum_{t=1}^{T}\sum_{i=1}^{n}(\overline{O}_i-\hat{y}_i)\frac{\partial O_i(t)}{\partial \mathbf{W}}.
    \label{eq:smse_W}
\end{equation}\normalsize
The derivative term $\frac{\partial O_i(t)}{\partial \mathbf{W}}$ in Eq. \ref{eq:sce_W} and Eq. \ref{eq:smse_W} can be calculated with the chain rule given by Eq. \ref{eq:stbp}.

\subsection{Temporal Efficient Training for SNN}
Temporal efficient training (TET) \citep{deng2022tet} is a direct training method for SNN, which minimizes the divergence between the output and target distribution at each time step. Mathematically, it can be described as
\begin{equation}
    \begin{split}
        \mathcal{L}_\mathrm{TET}=\frac{1}{T}\sum_{t=1}^{T}\biggl((1-\mu)\mathcal{L}_\mathrm{CE}\Big(O(t),y\Big)+\mu\mathcal{L}_\mathrm{MSE}\Big(O(t),\phi\Big)\biggr).
    \end{split}
    \label{eq:tet}
\end{equation}
In Eq. \ref{eq:tet} the term $\mathcal{L}_\mathrm{MSE}$ acts as a regularization component that constrains each time step's output in order to reduce the risk of outliers, $\phi$ is a constant used to regularize the membrane potential distribution and $\mu$ is a hyperparameter that controls the proportion of the regular term. \par
TET has the ability to compensate for the loss of momentum in the gradient descent process and to improve the temporal scalability of SNN, making it a great fundamental method for further improvement.

\subsection{Temporal Regularization Training}
Based on TET method, we propose a new kind of loss function $\mathcal{L}_\mathrm{TRT}$ to perform temporal regularization training (TRT). It can be described as
\begin{equation}
    r(t)=\sum_{i=1}^{l-1}\frac{\lambda}{1+(|\mathbf{W}_{i}|+\epsilon)\cdot(exp(\delta{\cdot}(t-1))-1)}\odot{\mathbf{W}^2_{i}}
    \label{eq:reg}
\end{equation}
\begin{equation}
    \begin{split}
        \mathcal{L}_\mathrm{TRT}=\frac{1}{T}\sum_{t=1}^{T}\biggl((1-\eta)\mathcal{L}_\mathrm{CE}\Big(O(t),y\Big)+\eta\mathcal{L}_\mathrm{MSE}\Big(O(t),\hat{y}\Big)+r(t)\biggr).
    \end{split}
    \label{eq:reg_loss}
\end{equation}
The core innovation of TRT lies in $r(t)$—a time-decaying regularizer that unleashes SNNs’ potential by forcing the model to learn robust features in early timesteps during training. As can be seen in \ref{eq:reg} , $l$ denotes the total number of layers in SNN, $\mathbf{W}_{i}$ denotes the synaptic weight between the $i^{th}$ and the ${i+1}^{th}$ layer of the network, $\delta$ is a temporal decay factor, and $\epsilon$ is a minimal value that safeguards the temporal decay term against zero-weight circumstances. The temporal regular term $r(t)$ exhibits time-dependent decay modulated by synaptic weights. In Eq. \ref{eq:reg_loss}, $\lambda$ is a factor that control the ratio of the term $r(t)$. As shown in Eq. \ref{eq:sce}, the cross-entropy loss drives the model to learn from the distribution of labels instead of truth values, which often leads to outliers when predicting challenging samples. We incorporate the $\mathcal{L}_\mathrm{MSE}$ term controlled by $\eta$ to constrain the predicted distribution, thereby reducing outliers and enhancing the model's performance. Experiments in the next part validate the effectiveness of this incorporation. Following the previous settings, $y$ still denotes the real label and $\hat{y}$ denotes the one-hot encoded label. The above-mentioned $\delta$, $\epsilon$, $\eta$ and $\lambda$ are implemented as hyperparameters.\par
Accroding to Eq. \ref{eq:lif_charge} and Eq. \ref{eq:stbp}, the gradient of weight $\mathbf{W}_{i}$ under the loss function $\mathcal{L}$ can be calculated as. The gradient of weight between the $i^{th}$ and the ${(i+1)}^{th}$ under the loss function $\mathcal{L}_\mathrm{TRT}$ can be calculated as
\begin{equation}
    \begin{split}%
        \frac{\partial \mathcal{L}_\mathrm{TRT}}{\partial \mathbf{W}_i}=\frac{1}{T}\sum_{t=1}^{T}\biggl((1-\eta)\frac{\partial \mathcal{L}_\mathrm{CE}(O(t),y)}{\partial u_i(t)}\frac{\partial u_i(t)}{\partial \mathbf{W}_i} \\
        +\eta\frac{\partial \mathcal{L}_\mathrm{MSE}(O(t),\hat{y})}{\partial u_i(t)}\frac{\partial u_i(t)}{\partial \mathbf{W}_i}+\frac{\partial r(t)}{\partial \mathbf{W}_i}\biggr).
    \end{split}
    \label{eq:trt_gradient}
\end{equation}
Following the analysis performed by \citet{wu2018spatio} and \citet{meng2023bptt}, the gradient of weight $\mathbf{W}_{i}$ under any loss functions implemented in $\mathcal{L}_\mathrm{TRT}$ can be calculated as
\begin{equation}
    \begin{split}
        \frac{\partial \mathcal{L}}{\partial \mathbf{W_i}} & =\frac{1}{T}\sum_{t=1}^{T}\frac{\partial \mathcal{L}}{\partial u_i(t)}\frac{\partial u_i(t)}{\partial \mathbf{W}_i}\\
        & =\frac{1}{T}\sum_{t=1}^{T}\biggl(\nabla_{\mathbf{W}_i}\mathcal{P}(t)+\nabla_{\mathbf{W}_i}\mathcal{T}(t)\biggr)\frac{\partial u_i(t)}{\partial \mathbf{W}_i},
    \end{split}
\end{equation}
where $\nabla_{\mathbf{W}_i}\mathcal{P}(t)$ and $\nabla_{\mathbf{W}_i}\mathcal{T}(t)$ can be further expanded as
\begin{equation}
    \nabla_{\mathbf{W}_i}\mathcal{P}(t)=\frac{\partial \mathcal{L}}{\partial s_i(t)}\frac{\partial s_i(t)}{\partial u_i(t)},
\end{equation}
\begin{equation}
    \nabla_{\mathbf{W}_i}\mathcal{T}(t)=\frac{1}{T}\sum_{t'=t+1}^{T}\biggl(\frac{\partial \mathcal{L}}{\partial s_i(t')}\frac{\partial s_i(t')}{\partial u_i(t')}\prod_{t''=1}^{t'-t}\xi_i(t'-t'')\biggr),
\end{equation}
where $t'\in[t+1,T]$, and $\xi_i(t)$ can be defined as
\begin{equation}
    \begin{split}
        \xi_i(t) & =\frac{\partial u_i(t+1)}{\partial u_i(t)}+\frac{\partial u_i(t+1)}{\partial s_i(t)}\frac{\partial s_i(t)}{\partial u_i(t)} \\
                 & =\gamma\big(1-s_i(t)-u_i(t)\mathbb{H}[u_i(t)]\big),
    \end{split}
    \label{eq:xi}
\end{equation}
where term $\mathbb{H}(\cdot)$ denotes the triangle-shaped surrogate gradient we used in this work.
In this case, the $\nabla_{\mathbf{W}_i}\mathcal{T}(t)$ can be rewrite as
\begin{equation}
    \nabla_{\mathbf{W}_i}\mathcal{T}(t)=\frac{1}{T}\sum_{t'=t+1}^{T}\biggl(\frac{\partial \mathcal{L}}{\partial s_i(t')}\frac{\partial s_i(t')}{\partial u_i(t')}\gamma^{(t'-t)}\prod_{t''=1}^{t'-t}\big(1-s_i(t'-t'')-u_i(t'-t'')\mathbb{H}(u_i[t'-t''])\big)\biggr).
\end{equation}
For different layers of the network, we have
\begin{equation}
  \begin{aligned}
    \frac{\partial \mathcal{L}}{\partial s_i(t)} =
    \begin{cases}
      \frac{\partial \mathcal{L}}{\partial s_i(t)},  & i=l \\
      \frac{\partial \mathcal{L}}{\partial u_{i+1}(t)}\frac{\partial u_{i+1}(t)}{\partial s_i(t)}, & i=l-1,\dots 1 \\
    \end{cases} \\
  \end{aligned}.
\end{equation}
As can be seen in Eq. \ref{eq:xi} and Eq. \ref{eq:lif_charge}, the term $\gamma$ is a time constant that controls the decay of the membrane potential, and is typically set to a value that smaller than 1 (e.g. $\gamma=0.5$ \citep{ding2024shrinking}). In this case, $\xi_i(t)$ eventually converges to 0 after continuous multiplication. This can be easily proved in the following process.\par
We reconstruct Eq. \ref{eq:xi} to $f_{\xi_i}(n)=\prod_{t=1}^{n}\xi_i(t)$, then we have
\begin{equation}
    \begin{split}
        \lim_{n \to \infty}f_\xi(n)
         & =\lim_{n \to \infty}\prod_{t=1}^{n}\xi_i(t)                                             \\
         & =\lim_{n \to \infty}\prod_{t=1}^{n}\gamma\big(1-s_i(t)-u_i(t)\mathbb{H}[u_i(t)]\big)    \\
         & =\lim_{n \to \infty}\gamma^n\prod_{t=1}^{n}\big(1-s_i(t)-u_i(t)\mathbb{H}[u_i(t)]\big), \\
    \end{split}
    \label{eq:proof}
\end{equation}
where $\gamma \in (0,1)$, resulting in $\lim_{n \to \infty}\gamma^n=0$.\par
From Eq. \ref{eq:proof}, we can obtain that $\prod_{t''=1}^{t'-t}\xi_i(t)$ eventually converge to 0 as $t'-t$ increases. This causes the value of $\nabla_{\mathbf{W}_i}\mathcal{T}(t)$ to also converge to 0 in early time steps, resulting in temporal gradient vanishing and limiting the performance of SNNs trained by the SDT method.\par
Our proposed TRT method implement the time-decaying regular term $r(t)$ into the loss function. During the training process, the gradient of $r(t)$ with respect to the synaptic weight $\mathbf{W}_{i,i+1}$ can be calculated as:
\begin{equation}
    \begin{split}
        \frac{\partial r(t)}{\partial \mathbf{W}_{i}}=\lambda\biggl[\frac{2\mathbf{W}_{i}}{1+(|\mathbf{W}_{i}|+\epsilon)(exp(\delta(t-1))-1)} \\
            -\frac{\mathbf{W}_{i}^2\odot\text{sign}(\mathbf{W}_{i})\odot(exp(\delta(t-1))-1)}{[1+(|\mathbf{W}_{i}|+\epsilon)(exp(\delta(t-1))-1)]^2}\biggr],
    \end{split}
    \label{eq:dr}
\end{equation}
where the gradient naturally decreases as the timestep $t$ increases, making $r(t)$ a temporal regular term that decays with time. As can be seen in Eq.\ref{eq:trt_gradient}, adding $\frac{\partial r(t)}{\partial \mathbf{W}_{i}}$ helps to compensate the temporal gradient of term $\mathcal{L}_\mathrm{CE}$ and $\mathcal{L}_\mathrm{MSE}$ in early timesteps who suffer from severe temporal gradient vanishing issue.\par
Like other similar approaches, TRT is applied exclusively during the training phase of the SNN, while the inference rules remain unchanged. The algorithm is detailed in Algorithm \ref{alg:trt}.
\begin{algorithm}[!ht]
    \caption{Temporal regularization training for single epoch}
    \KwIn{The SNN model; Simulation length: $T$; Training data; Number of training batches in one epoch: $B_{train}$; Regularization factor: $\lambda$; Ratio control factor: $\eta$; Time decay factor: $\delta$; Safeguard value: $\epsilon$.}
    \For{$i=1,\dots,B_{train}$}{
    Get training data, and class label: $\mathbf{Y}^i$;\\
    Use $\mathbf{Y}^i$ to compute one-hot encoded label: $\hat{\mathbf{Y}}^i$;\\
    Compute the SNN output $\mathbf{O}^i(t)$ of each time step;\\
    Calculate regular term $\mathbf{R}(t)$ for SNN output on each time step: $\mathbf{R}^i(t)=\sum_{j=1}^{l-1}\frac{\lambda}{1+(|\mathbf{W}_{j}|+\epsilon)\cdot(exp(\delta{\cdot}(t-1))-1)}\odot(\mathbf{W}_{j})^2$;\\
    Calculate the loss of SNN model in training epoch: $\mathcal{L}_\mathrm{TRT}=\frac{1}{T}\sum_{t=1}^{T}\biggl((1-\eta)\mathcal{L}_\mathrm{CE}\Big(\mathbf{O}^i(t),\mathbf{Y}^i\Big)+\eta\mathcal{L}_\mathrm{MSE}\Big(\mathbf{O}^i(t),\hat{\mathbf{Y}}^i\Big)+\mathbf{R}^i(t)\biggr)$;\\
    Backpropagation and update model parameters.
    }
    \label{alg:trt}
\end{algorithm}

\begin{table}[!ht]
    \centering
    \captionsetup{font=normalsize,justification=justified,singlelinecheck=false,width=\textwidth}
    \caption{Comparison with existing methods. We compare our method with the results reported in the respective literature. T denotes the simulation length.}
    \begin{threeparttable}
        {\scriptsize
            \begin{tabularx}{\textwidth}{
                    >{\centering\hsize=0.8\hsize}C
                    >{\centering\hsize=1.5\hsize}C
                    C
                    >{\centering\hsize=0.6\hsize}C
                    C}
                \toprule
                \textbf{Dataset} & \textbf{Methods}                                  & \textbf{Architecture} & \textbf{T}           & \textbf{Accuracy}         \\
                \toprule
                \multirow{7.5}{*}{\textbf{CIFAR10}}
                                 & STBP-tdBN (\citeyear{zheng2021tdbn})              & ResNet-19             & 6, 4, 2              & $93.16$, $92.92$, $92.34$ \\
                \cmidrule{2-5}
                                 & TET (\citeyear{deng2022tet})                      & ResNet-19             & 6, 4, 2              & $94.50$, $94.44$, $94.16$ \\
                \cmidrule{2-5}
                                 & TAB (\citeyear{jiang2024tab})                     & ResNet-19             & 6, 4, 2              & $94.81$, $94.76$, $94.73$ \\
                \cmidrule{2-5}
                                 & SLT-TET (\citeyear{anumasa2024slt})               & ResNet-19             & 6, 4, 2              & $95.26$, $95.18$, $94.96$ \\
                \cmidrule{2-5}
                                 & \textbf{TRT}                                      & \textbf{ResNet-19}    & \makecell{\textbf{6}                             \\\textbf{4}\\\textbf{2}} & \makecell{$\mathbf{95.72}\pm{0.40}$\\ $\mathbf{95.58}\pm{0.10}$\\ $\mathbf{95.30}\pm{0.23}$}\\
                \midrule
                \multirow{8.5}{*}{\textbf{CIFAR100}}
                                 & STBP-tdBN (\citeyear{zheng2021tdbn})              & ResNet-19             & 6, 4, 2              & $71.12$, $70.86$, $69.41$ \\
                \cmidrule{2-5}
                                 & TET (\citeyear{deng2022tet})                      & ResNet-19             & 6, 4, 2              & $74.72$, $74.47$, $72.87$ \\
                \cmidrule{2-5}
                                 & TAB (\citeyear{jiang2024tab})                     & ResNet-19             & 6, 4, 2              & $76.82$, $76.81$, $76.31$ \\
                \cmidrule{2-5}
                                 & SLT-TET (\citeyear{anumasa2024slt})               & ResNet-19             & 6, 4, 2              & $74.87$, $75.01$, $73.77$ \\
                \cmidrule{2-5}
                                 & \textbf{TRT}                                      & \textbf{ResNet-19}    & \makecell{\textbf{6}                             \\\textbf{4}\\\textbf{2}} & \makecell{$\mathbf{79.66}\pm{0.48}$\\ $\mathbf{79.51}\pm{0.34}$\\ $\mathbf{77.94}\pm{0.33}$}\\
                \midrule
                \multirow{5}{*}{\textbf{ImageNet100}}
                                 & EfficientLIF-Net (\citeyear{kim2023efficientlif}) & ResNet-19             & 5                    & $79.44$                   \\
                \cmidrule{2-5}
                                 & TET+LocalZO (\citeyear{mukhoty2023localzo})       & SEW-ResNet34          & 4                    & $78.58$                   \\
                \cmidrule{2-5}
                                 & IMP+LTS (\citeyear{shen2024implts})               & SEW-ResNet18          & 4                    & $80.80$                   \\
                \cmidrule{2-5}
                                 & \textbf{TRT}                                      & \textbf{SEW-ResNet34} & \textbf{4}           & $\mathbf{81.03}\pm{0.38}$ \\
                \bottomrule
            \end{tabularx}
        }
    \end{threeparttable}
    \label{tab:comparison_static}
\end{table}

\section{Experiments}
In this section, we demonstrate the effectiveness of our proposed TRT method and reveal the mechanism of TRT. We first compare our results with other existing state-of-the-art methods, then carry out ablation studies and analyzes to evaluate different aspects of our proposed TRT method. Furthermore, we establish a theoretical interpretation of its temporal regularization mechanism based on the results of Fisher information analysis.
\subsection{Comparison with Existing Works}
We evaluate the accuracy performance of TRT and compare it with existing works on both static and neuromorphic datasets to validate its effectiveness. The network architectures in this work include ResNet-19 \citep{zheng2021tdbn}, VGGSNN \citep{deng2022tet}, and SEW-ResNet-34 \citep{fang2021sew}.
The detailed training hyperparameters for each dataset are listed in the \ref{sec:ap_training_details}.
\paragraph{\textbf{Static datasets}} We choose CIFAR \citep{krizhevsky2009cifar} and ImageNet100 \citep{deng2009imagenet} to test our TRT method. The CIFAR dataset actually contains two widely used image classification dataset: CIFAR10 and CIFAR100, which have the number of classes mentioned in their name. ImageNet100 \citep{deng2009imagenet} is a 100-class subset of ImageNet. For CIFAR10 and CIFAR100, we apply TRT on the ResNet-19 networks and run three times to report the mean and standard deviation.
As shown in Table \ref{tab:comparison_static}, TRT achieves the highest accuracy among all existing approaches on CIFAR10. Even in the simulation length setting $T=2$, our TRT method still has increments of $2.14\%$ and $1.22\%$ compared to STBP-tdBN (standard direct training) and TET with $T=6$, respectively. On CIFAR100, TRT can also achieve much better performance than STBP-tdBN and TET. In addition, we compare our method with other state-of-the-art methods \citep{jiang2024tab,anumasa2024slt}. As listed in Table \ref{tab:comparison_static}, our method TRT outperforms all state-of-the-art methods across across three simulation length settings. For ImageNet100, TRT is applied to a SEW-ResNet-34 network, which obtains $81.03\%$ accuracy. This result outperforms the state-of-the-art method and even reaches a competitive level with the performance of LocalZO \citep{mukhoty2023localzo} that utilized ImageNet policy for additional data augmentation ($81.56\%$).
\begin{table}[!ht]
    \centering
    \captionsetup{font=normalsize,justification=justified,singlelinecheck=false,width=\textwidth}
    \caption{Comparison with existing methods. We compare our method with the results reported in the respective literature. T denotes the simulation length. The symbol (*) denotes our implementation.}
    \begin{threeparttable}
        {\scriptsize
            \begin{tabularx}{\textwidth}{
                    >{\centering\hsize=1.1\hsize}C
                    >{\centering\hsize=1.5\hsize}C
                    C
                    >{\centering\hsize=0.3\hsize}C
                    C}
                \toprule
                \textbf{Dataset} & \textbf{Methods}                            & \textbf{Architecture} & \textbf{T}  & \textbf{Accuracy}         \\
                \toprule
                \multirow{5}{*}{\textbf{DVS-CIFAR10}}
                                 & SEW-ResNet (\citeyear{fang2021sew})         & 7B-wideNet            & 16          & $74.40$                   \\
                \cmidrule{2-5}
                                 & TET (\citeyear{deng2022tet})                & VGGSNN                & 10          & $  83.10$\tnote{*}        \\
                \cmidrule{2-5}
                                 & TET+LocalZO (\citeyear{mukhoty2023localzo}) & VGGSNN                & 10          & $81.87$                   \\
                \cmidrule{2-5}
                                 & \textbf{TRT}                                & \textbf{VGGSNN}       & \textbf{10} & $\mathbf{83.20}\pm{0.49}$ \\
                \midrule
                \multirow{5}{*}{\textbf{N-Caltech101}}
                                 & TET+LocalZO (\citeyear{mukhoty2023localzo}) & VGGSNN                & 10          & $82.99$                   \\
                \cmidrule{2-5}
                                 & TKS (\citeyear{dong2024temporal})           & VGG-TKS               & 10          & $84.10$                   \\
                \cmidrule{2-5}
                                 & IMP+TET-S (\citeyear{shen2024implts})       & VGGSNN                & 10          & $85.01$                   \\
                \cmidrule{2-5}
                                 & \textbf{TRT}                                & \textbf{VGGSNN}       & \textbf{10} & $\mathbf{85.18}\pm{0.20}$ \\
                \bottomrule
            \end{tabularx}
        }
    \end{threeparttable}
    \label{tab:comparison_neuromorphic}
\end{table}
\paragraph{\textbf{Neuromorphic datasets}} We apply the TRT method to a simple spiking neural network VGGSNN and compare the results with state-of-the-art methods on neuromorphic datasets. We choose DVS-CIFAR10 \citep{li2017dvscifar10} and N-Caltech101 \citep{orchard2015converting} to test our proposed method. DVS-CIFAR10 is a 10-class dataset that is usually considered as the most challenging mainstream neuromorphic dataset. N-Caltech101 is a 101-class dataset converted from the well-known static dataset Caltech101 \citep{fei2004caltech101}. For DVS-CIFAR10 and N-Caltech101, we reduce the spatial resolution to 48$\times$48, split these two datasets in a $9:1$ ratio, and report the mean and standard deviation.
As shown in Table \ref{tab:comparison_neuromorphic}, our proposed TRT method obtains $83.20\%$ accuracy on DVS-CIFAR10 and $85.14\%$ accuracy on N-Caltech101, which outperforms most existing approaches.
\subsection{Ablation Study}
\begin{figure}[!ht]
    \centering
    \includegraphics[width=0.6\textwidth]{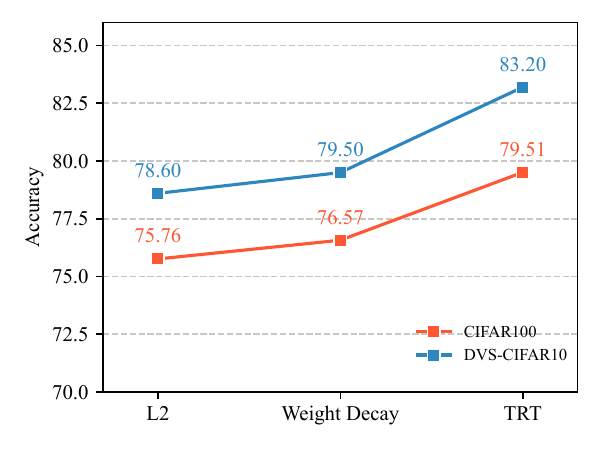}
    \captionsetup{font=normalsize,justification=justified,singlelinecheck=false,width=\columnwidth}
    \caption{The accuracy results of different regularization methods on two datasets. For CIFAR100, we set simulation length $T$ to 4, then apply L2 regularization and weight decay both with the value of $2e-5$. For DVS-CIFAR10, we set $T$ to 10, then apply L2 regularization and weight decay both with the value of $4e-5$.}
    \label{fig:different_regularization_methods}
\end{figure}

\paragraph{\textbf{Effectiveness of TRT}} We compare the results training with standard direct training (SDT) and TRT on SNNs with surrogate gradient. As shown in Table \ref{tab:trt_effectiveness}, our proposed TRT training method outperforms the SDT method on CIFAR10 dataset. TRT increases the accuracy by $2.56\%$, $2.66\%$, and $2.96\%$ when the simulation length is 2, 4, 6, respectively. On CIFAR100, TRT dramatically increases accuracy by $8.54\%$, $8.65\%$, and $8.53\%$. On DVS-CIFAR10, TRT outperforms SDT by $4.70\%$. We further compare the results of TRT with L2 regularization and weight decay on CIFAR100 and DVS-CIFAR10. The results are summarized in Figure. \ref{fig:different_regularization_methods}. As shown in Figure. \ref{fig:different_regularization_methods}, TRT outperforms L2 regularization and weight decay in terms of accuracy, indicating the effectiveness of this method.
\begin{table}[!t]
    \centering
    \caption{Comparison between SDT and TRT. We adopt ResNet-19 on CIFAR10/100 and VGGSNN on DVS-CIFAR10. The symbol (*) denotes our implementation.}
    \begin{threeparttable}
        {\scriptsize
            \begin{tabularx}{\columnwidth}{
                    P{0.25\hsize}
                    C
                    P{0.1\hsize}
                    P{0.15\hsize}
                    P{0.15\hsize}}
                \toprule
                                                     &                                           &                                & \multicolumn{2}{c}{\textbf{Method}}                    \\
                \cmidrule(l){4-5}
                \multirow{-2.5}{*}{\textbf{Dataset}} & \multirow{-2.5}{*}{\textbf{Architecture}} & \multirow{-2.5}{*}{\textbf{T}} & \textbf{SDT}                        & \textbf{TRT}     \\
                \toprule
                \textbf{CIFAR10}                     & ResNet-19                                 & \makecell{6                                                                             \\4\\2} & \makecell{$93.16$\\ $92.92$\\ $92.34$} & \makecell{$\mathbf{95.72}$\\ $\mathbf{95.58}$\\ $\mathbf{95.30}$}\\
                \midrule
                \textbf{CIFAR100}                    & ResNet-19                                 & \makecell{6                                                                             \\4\\2} & \makecell{$71.12$\\ $70.86$\\ $69.41$} & \makecell{$\mathbf{79.66}$\\ $\mathbf{79.51}$\\ $\mathbf{77.94}$}\\
                \midrule
                \textbf{DVS-CIFAR10}                 & VGGSNN                                    & 10                             & $78.50$\tnote{*}                    & $\mathbf{83.20}$ \\
                \bottomrule
            \end{tabularx}
        }
    \end{threeparttable}
    \label{tab:trt_effectiveness}
\end{table}
\begin{table}[!t]
    \centering
    \caption{Comparison of TRT accuracy results on CIFAR100 under different values of $\delta$.}
    \begin{threeparttable}
        {\scriptsize
            \begin{tabularx}{\columnwidth}{CCCCC}
                \toprule
                $\delta$ & $0$   & $\mathbf{0.25}$ & $0.5$ & $1.0$ \\
                \toprule
                Accuracy & 78.55 & \textbf{79.51}  & 78.99 & 79.37 \\
                \bottomrule
            \end{tabularx}
        }
    \end{threeparttable}
    \label{tab:different_delta}
\end{table}
\begin{figure}[!t]
    \centering
    \includegraphics[width=\columnwidth]{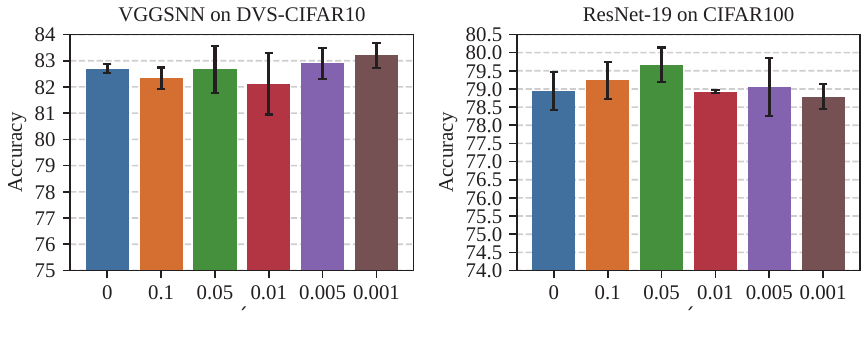}
    \captionsetup{font=normalsize,justification=justified,singlelinecheck=false,width=\columnwidth}
    \caption{The accuracy results under different values of $\eta$}
    \label{fig:different_eta}
\end{figure}
\paragraph{\textbf{Effect of different loss ratios}} Our proposed TRT training method uses a hyperparameter $\eta$ to control the ratio of $\mathcal{L}_\mathrm{CE}$ and $\mathcal{L}_\mathrm{MSE}$. In this section, we study the effects of different $\eta$ values. We examine the effect of 6 different values of $\eta$ on CIFAR100 and DVS-CIFAR10. The results are summarized in Figure. \ref{fig:different_eta}, which indicates that the combination of $\mathcal{L}_\mathrm{CE}$ and $\mathcal{L}_\mathrm{MSE}$ in a specific ratio leads to improved accuracy. As a result, we set $\eta$ to $5e-2$ for CIFAR10, CIFAR100 and N-Caltech101, $1e-3$ for DVS-CIFAR10 and ImageNet100.
\paragraph{\textbf{Effect of different decay rates on regular term}} The
proposed TRT method uses a hyperparameter $\delta$ to control the decay rate of the regular term $r(t)$ (Eq. \ref{eq:reg}). In this section, we compare the accuracy results under different values of $\delta$. We examine the effect of four different values of $\delta$ on CIFAR100. Table \ref{tab:different_delta} lists the experimental results. As a result, we set $\delta=0.25$ as a base value for general experiments (see \ref{sec:ap_supplemental_results}).

\begin{figure}[!t]
    \centering
    \begin{subfigure}[t]{0.31\textwidth}
        \centering
        \includegraphics[width=\linewidth]{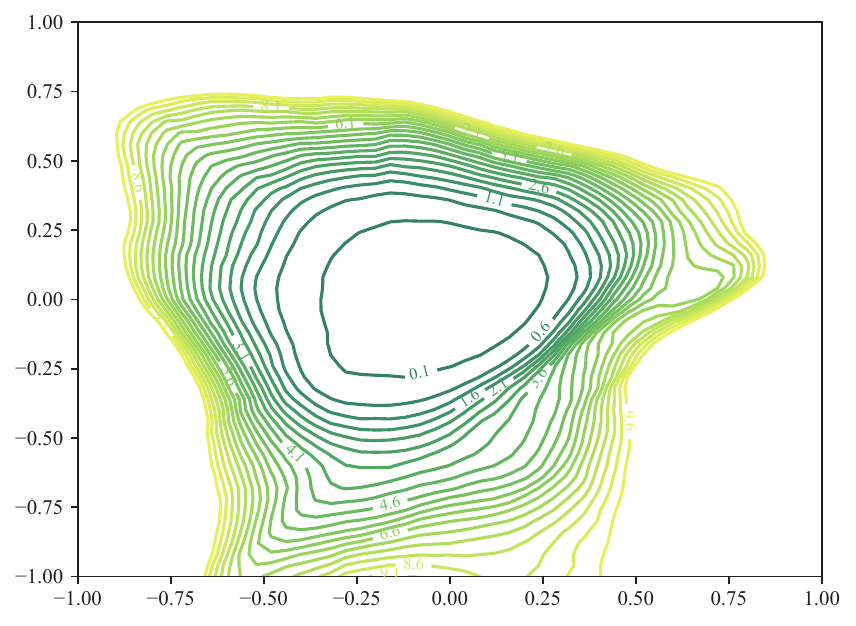}
        \caption{SDT, DVS-CIFAR10.}
        \label{fig:landscape_sdt_dvscifar10}
    \end{subfigure}
    \begin{subfigure}[t]{0.31\textwidth}
        \centering
        \includegraphics[width=\linewidth]{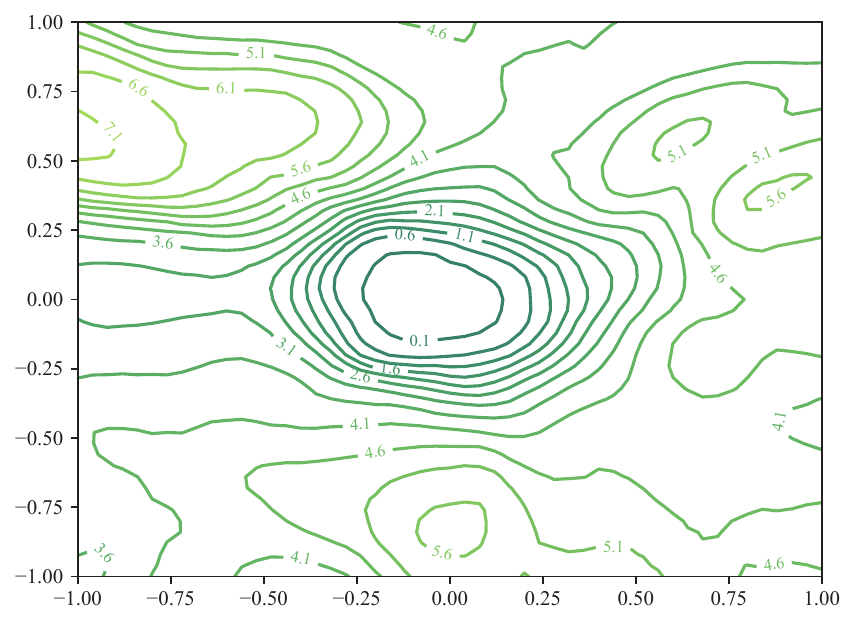}
        \caption{TET, DVS-CIFAR10.}
        \label{fig:landscape_tet_dvscifar10}
    \end{subfigure}
    \begin{subfigure}[t]{0.31\textwidth}
        \centering
        \includegraphics[width=\linewidth]{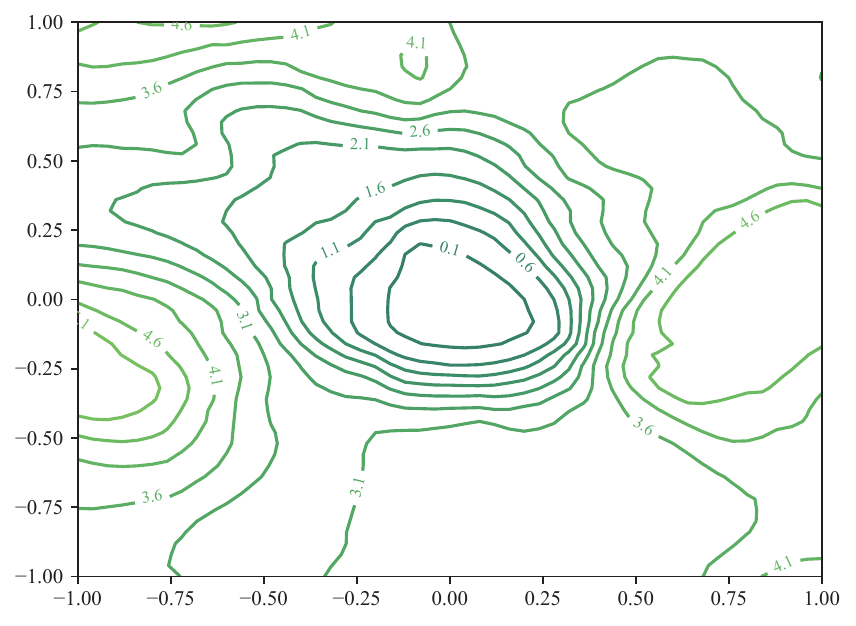}
        \caption{TRT, DVS-CIFAR10.}
        \label{fig:landscape_trt_dvscifar10}
    \end{subfigure}
    \begin{subfigure}[t]{0.31\textwidth}
        \centering
        \includegraphics[width=\linewidth]{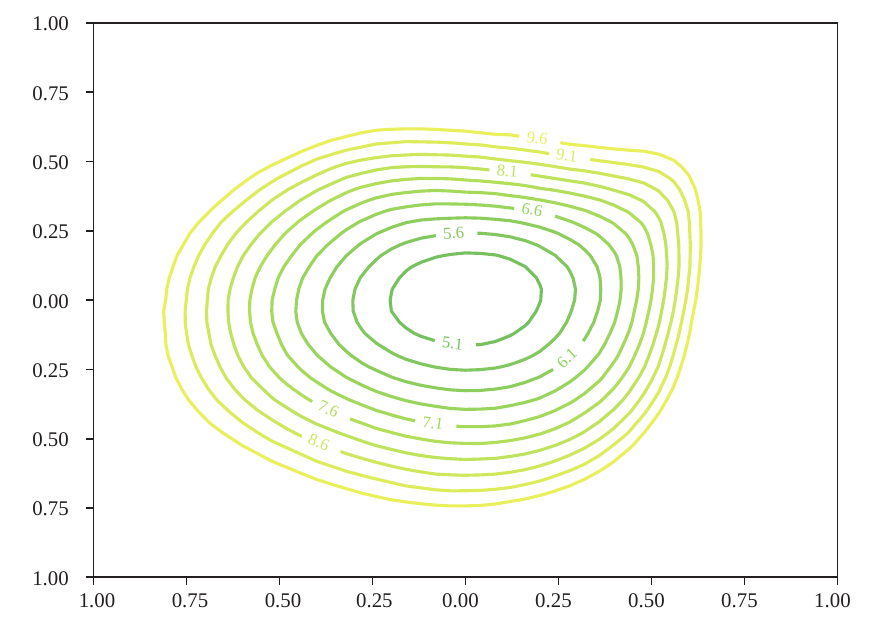}
        \caption{SDT, N-Caltech101.}
        \label{fig:landscape_sdt_ncaltech101}
    \end{subfigure}
    \begin{subfigure}[t]{0.31\textwidth}
        \centering
        \includegraphics[width=\linewidth]{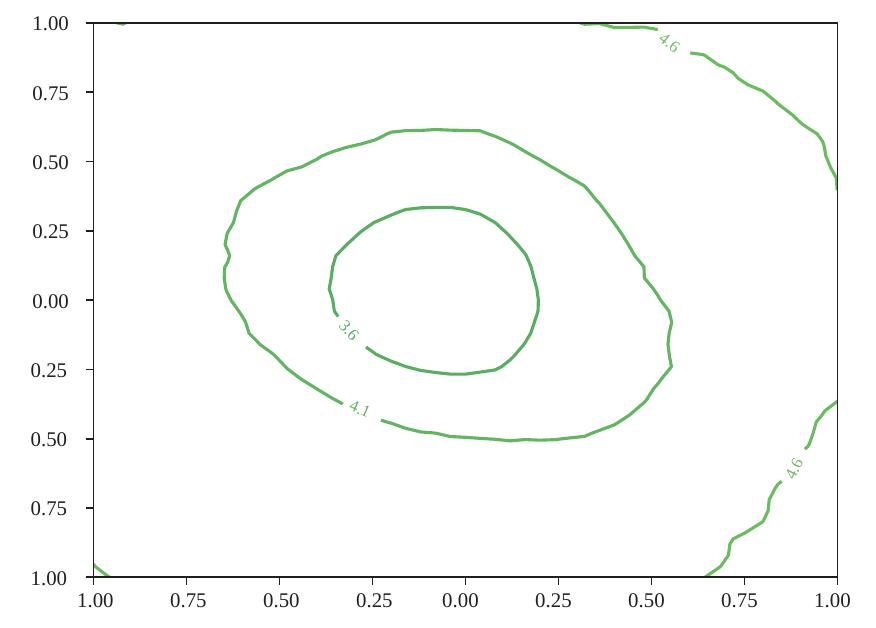}
        \caption{TET, N-Caltech101.}
        \label{fig:landscape_tet_ncaltech101}
    \end{subfigure}
    \begin{subfigure}[t]{0.31\textwidth}
        \centering
        \includegraphics[width=\linewidth]{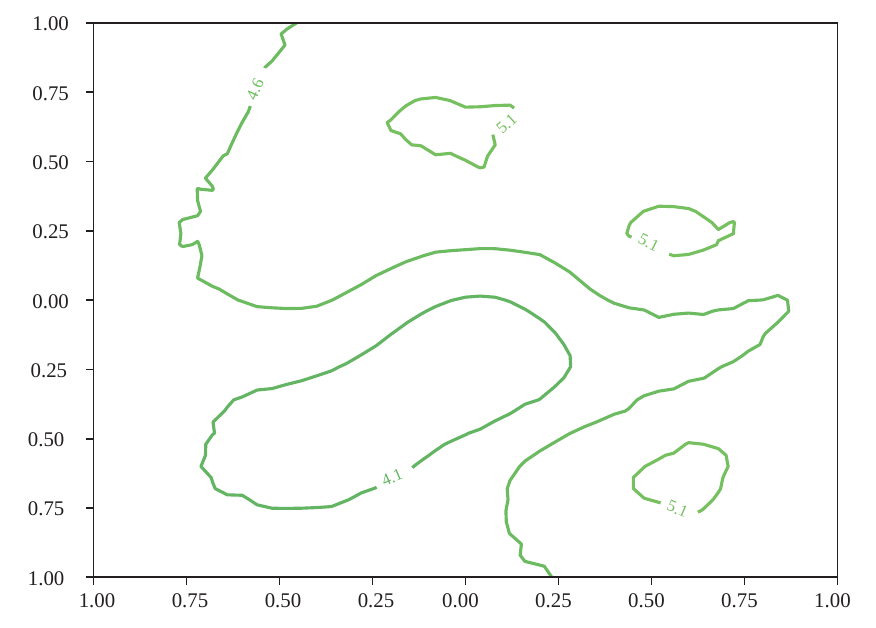}
        \caption{TRT, N-Caltech101.}
        \label{fig:landscape_trt_ncaltech101}
    \end{subfigure}
    \caption{Visualization of the loss landscapes of SDT, TET, and our TRT method on DVS-CIFAR10 and N-Caltech101 dataset.}
    \label{fig:landscape}
\end{figure}
\subsection{Analysis}
\paragraph{\textbf{Loss landscape visualization}} We further demonstrate the 2D loss landscapes \citep{li2018landscape} of TRT, TET, and SDT
on DVS-CIFAR10
and N-Caltech101 dataset to verify that our proposed TRT method provides SNNs with more generalizability (Figure. \ref{fig:landscape}). On DVS-CIFAR10, the area of TRT's local minima is slightly larger than TET. Although the SDT method exhibits a larger minimum region area in the loss landscape compared to TRT, the contour lines of TRT are sparser compared to SDT and TET. This indicates that TRT is flatter than SDT and TET in both the local minima area and the global scale. On N-Caltech101, the landscape of TRT is less flatter than TET on the global scale, but the area of TRT's local minima is significantly larger than TET and SDT, indicating that TRT helps the network to jump out of the sharp local minima area with poor generalization and find another local minima with flatter landscape. These results demonstrates that the TRT method exhibits smoother loss variations under parameter perturbations, indicating that TRT effectively improves the network generalization.

\begin{figure}[!t]
    \centering
    \begin{subfigure}[t]{0.325\textwidth}
        \centering
        \includegraphics[width=\columnwidth]{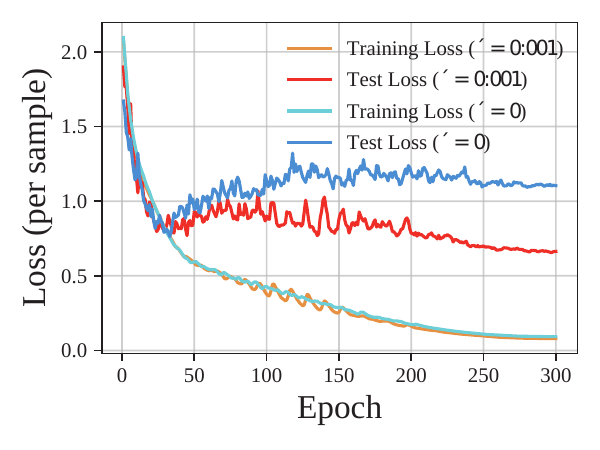}
        \caption{DVS-CIFAR10: TRT}
        \label{fig:learning_curves_trt_dvscifar10_merge}
    \end{subfigure}
    \begin{subfigure}[t]{0.325\textwidth}
        \centering
        \includegraphics[width=\columnwidth]{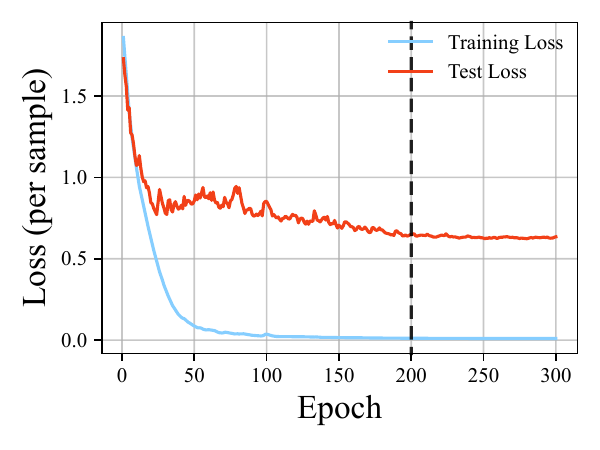}
        \caption{DVS-CIFAR10: TET.}
        \label{fig:learning_curves_tet_dvscifar10_content}
    \end{subfigure}
    \begin{subfigure}[t]{0.325\textwidth}
        \centering
        \includegraphics[width=\columnwidth]{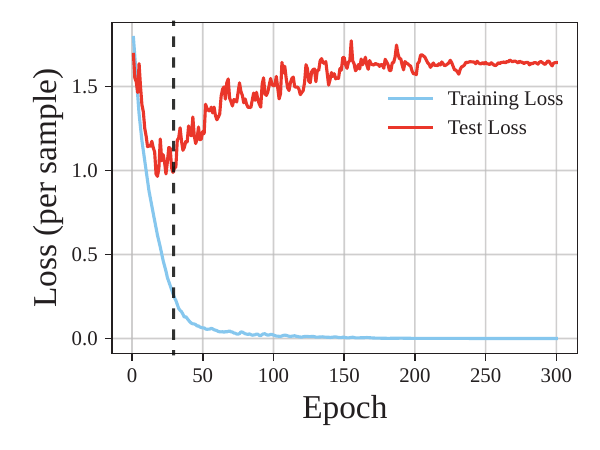}
        \caption{DVS-CIFAR10: SDT.}
        \label{fig:learning_curves_sdt}
    \end{subfigure}
    \begin{subfigure}[t]{0.325\textwidth}
        \centering
        \includegraphics[width=\columnwidth]{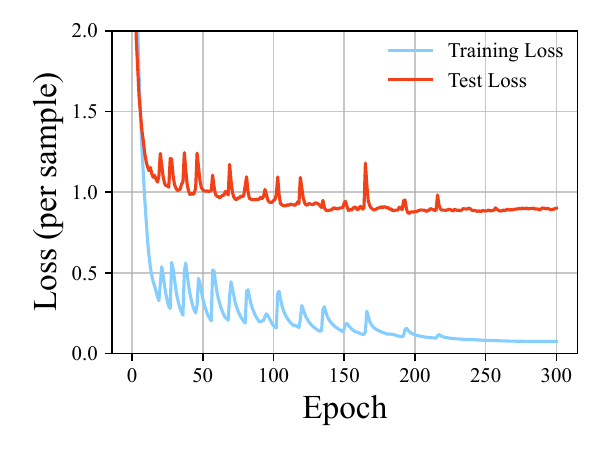}
        \caption{N-Caltech101: TRT.}
        \label{fig:learning_curves_trt_ncaltech101_content}
    \end{subfigure}
    \begin{subfigure}[t]{0.325\textwidth}
        \centering
        \includegraphics[width=\columnwidth]{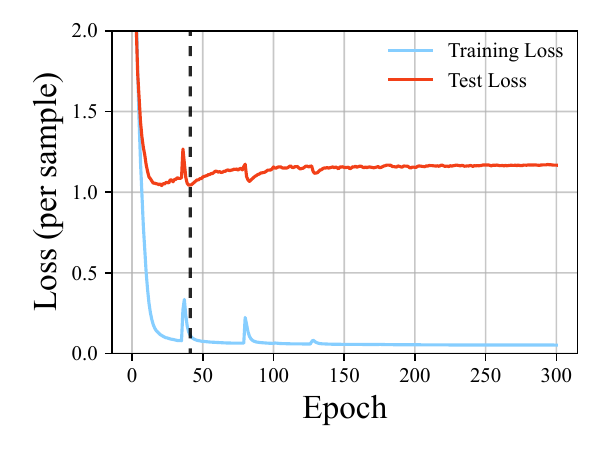}
        \caption{N-Caltech101: TET.}
        \label{fig:learning_curves_tet_ncaltech101_content}
    \end{subfigure}
    \begin{subfigure}[t]{0.325\textwidth}
        \centering
        \includegraphics[width=\columnwidth]{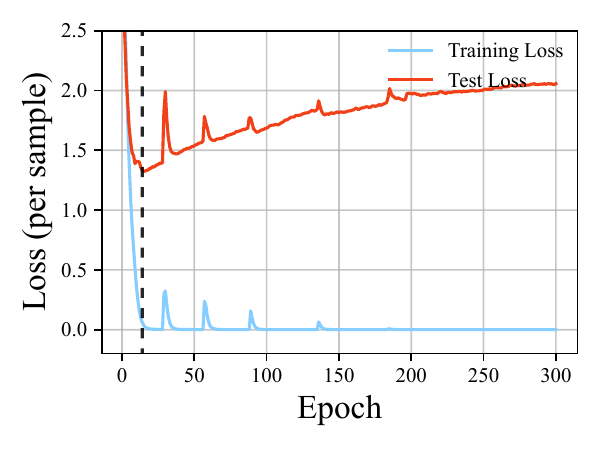}
        \caption{N-Caltech101: SDT.}
        \label{fig:learning_curves_sdt_ncaltech101_content}
    \end{subfigure}
    \caption{Learning curves on DVS-CIFAR10 and N-Caltech101. (a-c) depict the learning curves of TRT, TET, and SDT methods on DVS-CIFAR10, respectively. (d-f) depict the learning curves of TRT, TET, and SDT methods on N-Caltech101, respectively. Dash line denotes the 30th epoch. Dash lines denote the 200th, 30th, 45th and 20th epoch, respectively.}
    \label{fig:learning_curve_trt}
\end{figure}
\paragraph{\textbf{Effectiveness of TRT in mitigating overfitting}} In this section, we study the effectiveness of TRT in mitigating overfitting. We compare the learning curves of training VGGSNN on DVS-CIFAR10 dataset using the TRT method with different $\eta$ values. The criterion for the testing process is $\mathcal{L}_\mathrm{SCE}$. Figure. \ref{fig:learning_curve_trt} shows the curves of the TRT method with and without $\mathcal{L}_\mathrm{MSE}$ integration.
As indicated in Figure. \ref{fig:learning_curves_trt_dvscifar10_merge}, the training loss curves continue to decrease, while the test loss with $\eta=0.001$ shows a general downward trend, and the test loss with $\eta=0$ rapidly approaches stabilization. In contrast, TET shows a sharp initial drop in training loss in the early stages and then fluctuates at lower values, while the testing loss decreases slowly and remains between $0.5-0.7$ after the 200th epoch (Figure. \ref{fig:learning_curves_tet_dvscifar10_content}). As for the SDT method (Figure. \ref{fig:learning_curves_sdt}), the test loss increases significantly after the 30th epoch and stabilizes at an elevated plateau, while the training loss continues to decrease and stabilize around 0. On N-Caltech101, learning curves of these methods exhibit identical trends, which also proves TRT's ability on mitigating overfitting. Compared to TET and SDT, the trends of these learning curves indicate that, under the joint optimization of the term $\mathcal{L}_\mathrm{MSE}$ and the temporal regularization term $r(t)$, the
TRT method allows SNNs to mitigate overfitting and achieve superior generalization. 

\begin{figure}[!t]
    \centering
    \begin{subfigure}[t]{0.49\columnwidth}
        \centering
        \includegraphics[width=\textwidth]{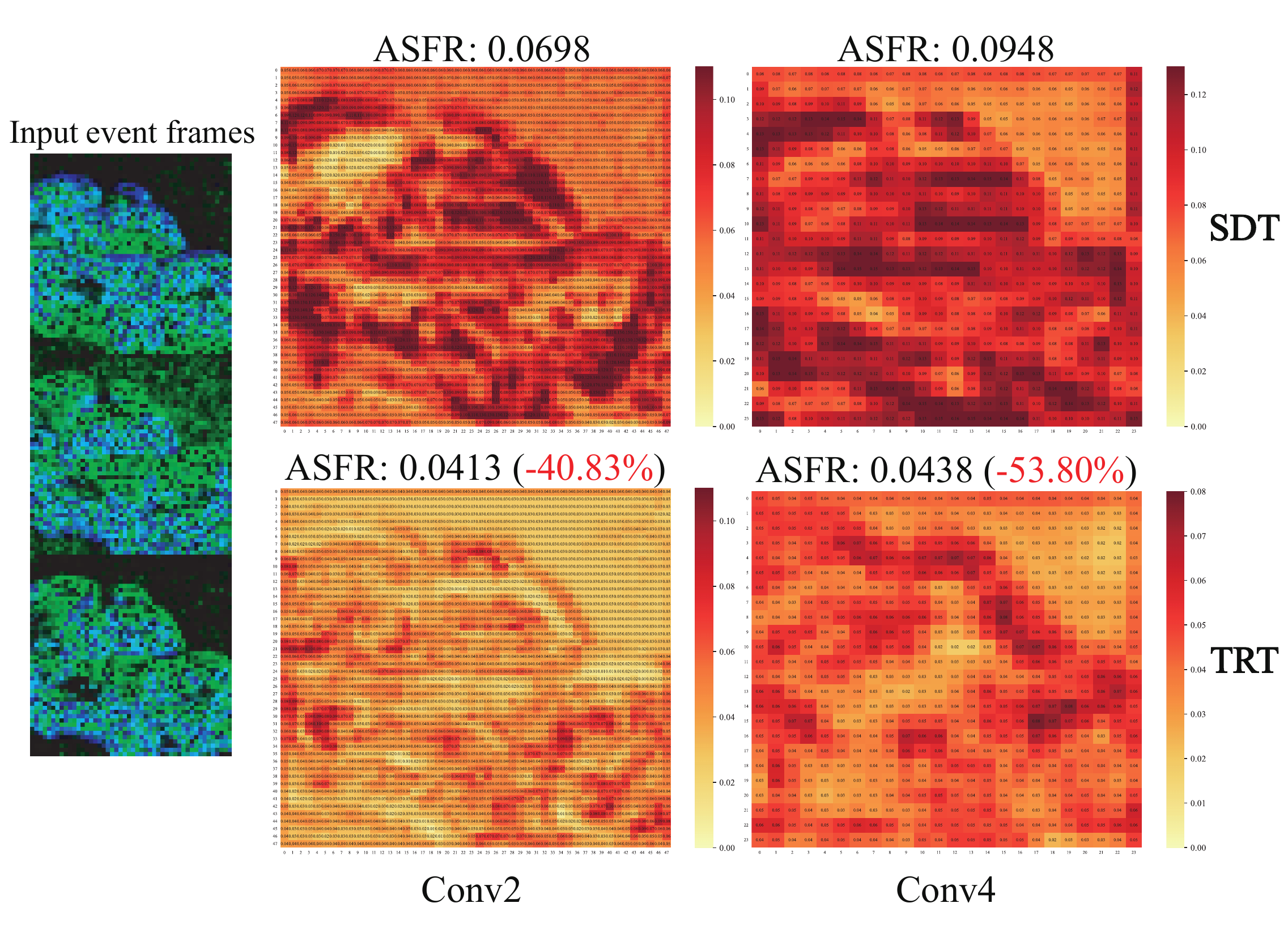}
        \caption{DVS-CIFAR10.}
        \label{fig:asfr_dvscifar10_small}
    \end{subfigure}
    \begin{subfigure}[t]{0.49\columnwidth}
        \centering
        \includegraphics[width=\textwidth]{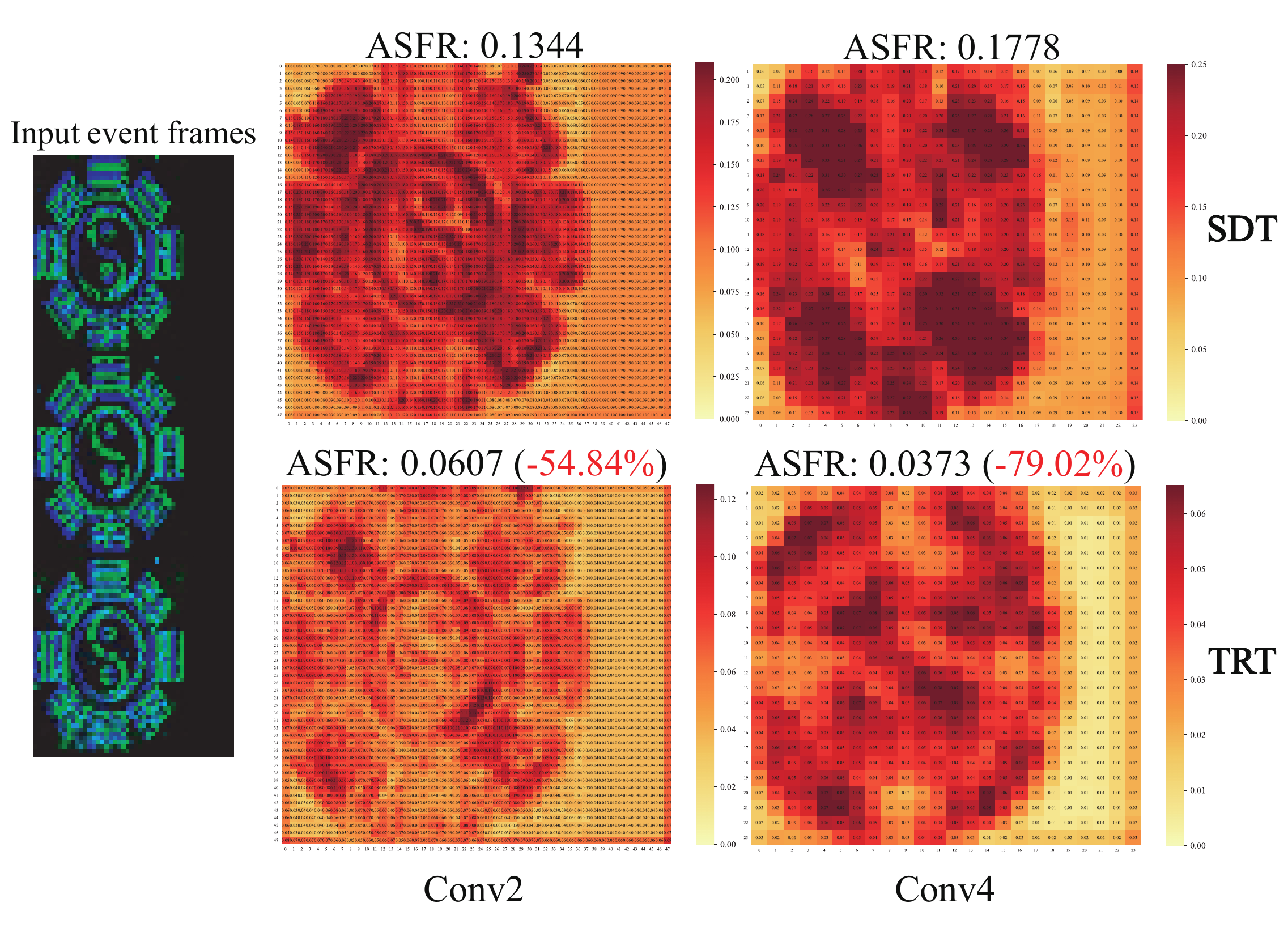}
        \caption{N-Caltech101.}
        \label{fig:asfr_ncaltech101_small}
    \end{subfigure}
    \caption{Visualization of ASFR in VGGSNN on DVS-CIFAR10 and N-Caltech101.}
    \label{fig:asfr_all}
\end{figure}

\paragraph{\textbf{Average spike firing rate visualization}} To verify the energy efficiency of TRT, we visualize the average spike firing rate (ASFR) in the first two layers of VGGSNN inferencing on DVS-CIFAR10 and N-Caltech101 dataset (Figure. \ref{fig:asfr_all}). Compared to the models trained from the SDT method, models trained from TRT achieve better performance with lower ASFR, indicating that TRT helps to train SNNs with low energy consumption, which are suitable for real-time application.

\begin{figure*}[!t]
    \centering
    \includegraphics[width=\textwidth]{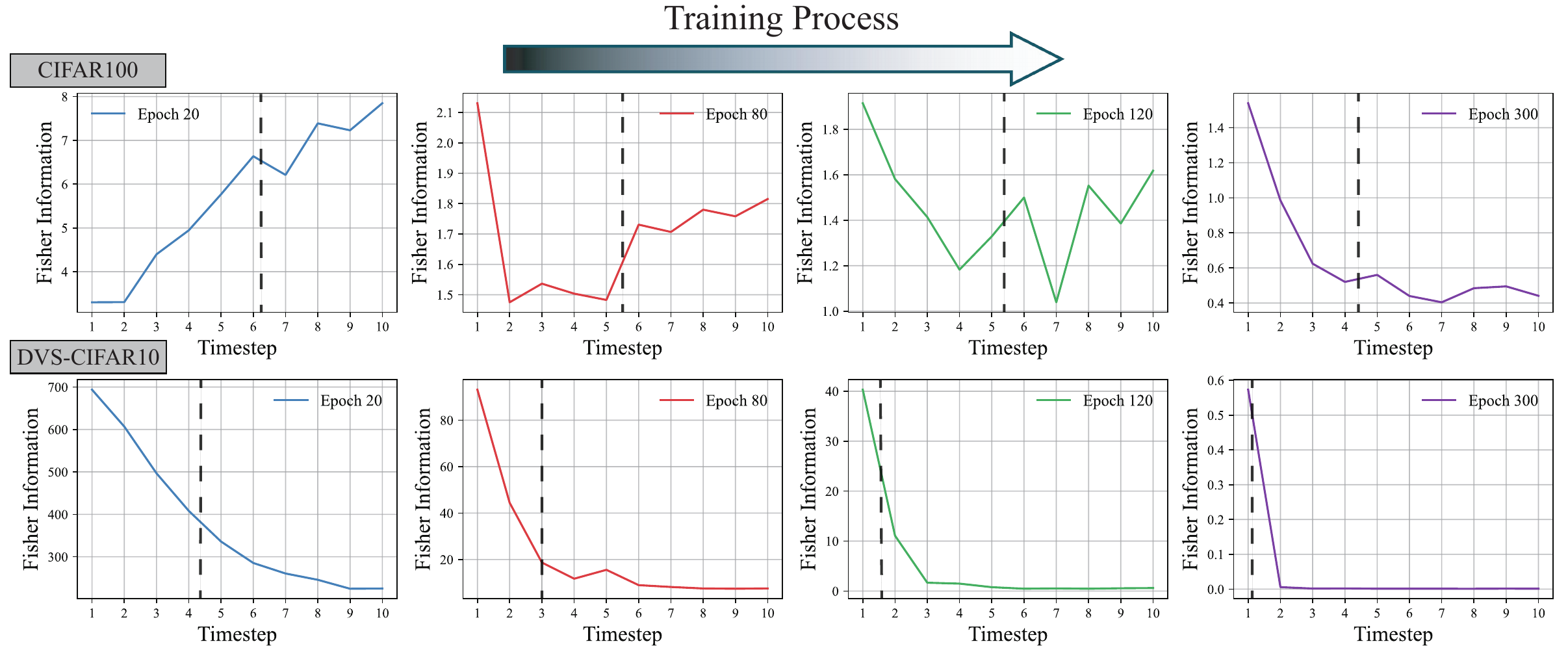}
    \caption{Illustration of temporal dynamics on CIFAR10 and DVS-CIFAR10. Dash line denotes the information centroid (IC).
    }
    \label{fig:fi_curves}
\end{figure*}
\subsection{Discussion}
As we mentioned before, different forms of temporal regularization have been applied to SNNs' training process in previous works, but still lack of reasonable explanations. In previous section, we analyzed how TRT enhances model performance by mitigating the vanishing temporal gradient issue in SNNs. Here we establish a theoretical interpretation for temporal regularization based on the TIC \citep{kim2023tic} phenomenon to elucidate its mechanism for improving model generalization. We measure the Fisher information of SNNs (ResNet-19 on CIFAR10 and VGGSNN on DVS-CIFAR10) across TRT training epochs (see Figure. \ref{fig:fi_curves}). The results for both datasets exhibit a significant temporal shift in information concentration, showing that the TRT training process does not change the temporal information dynamics within the SNN. In early training stages, Fisher information concentrates on the later timesteps. As training progresses, Fisher information shift from the latter timesteps to the early timesteps, indicating that in later training stages, the early timesteps gain greater importance and exert greater influence on model performance. This means that the presence of noise or outliers in early time steps can significantly disrupt model training in later stages, decreasing model generalization and predisposing the network to learn spurious features that exacerbate overfitting. Therefore, applying regularization that decays over time can constrain the synaptic weights of SNNs in the important early time steps more effectively, enhancing model generalization and mitigating overfitting, leading to higher performance.

\section{Conclusion}
In this paper, we propose the temporal regularization training algorithm (TRT), a direct training method that integrates temporal regularization. We perform theoretical analysis to reveal TRT's ability on mitigating temporal gradient vanishing. We validate the TRT method on various datasets, showing the state-of-the-art performance compared to other methods. Our findings confirm that TRT unleashes the latent generalization of SNNs, with experiments revealing flatter loss landscapes and smoother learning curves. Furthermore, we analyze the temporal information distribution within the SNN during the TRT training process. We empirically demonstrate that the Temporal Information Concentration (TIC) phenomenon, historically identified in previous studies, persistently emerges during the TRT process. According to the TIC phenomenon, we provide a reasonable explanation for the time-decaying regularization scheme in TRT and demonstrate that this interpretation generalizes to all time-decaying regularization methods.

\appendix
\setcounter{figure}{0}
\setcounter{table}{0}

\section{Experiment settings}
\label{sec:ap_training_details}
\subsection{Introduction to Datasets}
\begin{table}[!ht]
    \centering
    \caption{Overview of the datasets used in our experiments}
    \begin{threeparttable}
        {\scriptsize
            \begin{tabularx}{\columnwidth}{
                P{0.22\hsize}P{0.24\hsize}P{0.2\hsize}C}
                \toprule
                \textbf{Type} & \textbf{Dataset} & \textbf{Categories} & \textbf{Total samples} \\
                \toprule
                \multirow{4}{*}{\textbf{Static images}}
                              & CIFAR10          & 10                  & 60000                  \\
                \cmidrule{2-4}
                              & CIFAR100         & 100                 & 60000                  \\
                \cmidrule{2-4}
                              & ImageNet100      & 100                 & 135000                 \\
                \midrule
                \multirow{2.5}{*}{\textbf{Event data}}
                              & DVS-CIFAR10      & 10                  & 10000                  \\
                \cmidrule{2-4}
                              & N-Caltech101     & 101                 & 8709                   \\
                \bottomrule
            \end{tabularx}
        }
    \end{threeparttable}
    \label{tab:ap_dataset}
\end{table}
Table \ref{tab:ap_dataset} gives an overview of the datasets used in our experiments.
\paragraph{\textbf{CIFAR}} The CIFAR dataset \citep{krizhevsky2009cifar} contains two subsets: CIFAR10 and CIFAR100, each of them contains 60,000 static images of size 32$\times$32. In CIFAR10 the images are divided into 10 classes while in CIFAR100 there are 100 classes. For both datasets, 50,000 images are used for training and 10,000 for testing.
\paragraph{\textbf{ImageNet100}} ImageNet100 \citep{deng2009imagenet} is a subset of the widely used ImageNet. ImageNet100 contains 100 classes, each of them provides 1300 training samples and 50 testing samples.
\paragraph{\textbf{DVS-CIFAR10}} DVS-CIFAR10 \citep{li2017dvscifar10} is an event image dataset converted from CIFAR10. It contains 10,000 samples with the size 128$\times$128.
\paragraph{\textbf{N-Caltech101}} The N-Caltech101 dataset \citep{orchard2015converting} is a neuromorphic version converted from the Caltech101 dataset \citep{fei2004caltech101}. N-Caltech 101 removes the class "faces" from the original dataset with "simple faces". N-Caltech101 has 8709 samples with 100 object classes and 1 background class.

\subsection{Data Processing Methods}
\paragraph{\textbf{Static datasets}} For both CIFAR10 and CIFAR100, we apply cutout (one hole) and random horizontal flip and crop with data normalization to the training images. For ImageNet100, following the previous study \citep{he2016resnet}, 224$\times$224 crop and horizontal flip with data normalization are applied to training images, and 224$\times$224 resize and central crop with data normalization are applied to testing images.
\paragraph{\textbf{Neuromorphic datasets}} For both N-Caltech101 and DVS-CIFAR10, we divide the data stream into 10 blocks by time and accumulate spikes in each block. Both datasets are split into training and testing sets with a $9:1$ partitioning ratio, and then resized to 48$\times$48. For DVS-CIFAR10, random horizontal flip and random roll are applied to training data.

\begin{table*}[!ht]
    \centering
    \caption{Hyperparameter settings for experiments. The leaky factor $\gamma$ of membrane potential equals to $\frac{1}{\tau}$. Minimum learning rate of the scheduler is set to 0.}
    \begin{threeparttable}
        {\scriptsize
            \begin{tabularx}{\textwidth}{
                    >{\raggedright}C
                    CCCC}
                \toprule
                                    & \textbf{CIFAR10/100} & \textbf{ImageNet100} & \textbf{DVS-CIFAR10} & \textbf{N-Caltech101} \\
                \toprule
                Number epochs       & 300                  & 300                  & 300                  & 300                   \\
                Mini batch size     & 64                   & 64                   & 64                   & 64                    \\
                Learning Rate       & 1e-3                 & 1e-3                 & 1e-3                 & 1e-3                  \\
                T                   & 6,4,2                & 4                    & 10                   & 10                    \\
                Neuron: $\tau$      & 2.0                  & 1.0                  & 2.0                  & 2.0                   \\
                Neuron: $u_{th}$    & 1.0                  & 1.0                  & 1.0                  & 1.0                   \\
                Neuron: $u_{reset}$ & 0                    & 0                    & 0                    & 0                     \\
                Neuron: $\alpha$    & 1.0                  & 1.0                  & 1.0                  & 1.0                   \\
                TRT: $\eta$         & 0.05                 & 0.001                & 0.001                & 0.05                  \\
                TRT: $\lambda$      & 1e-5                 & 5e-5                 & 5e-5                 & 5e-5                  \\
                TRT: $\delta$       & 0.25                 & 0.5                  & 0.5                  & 0.5                   \\
                TRT: $\epsilon$     & 1e-5                 & 1e-5                 & 1e-5                 & 1e-5                  \\
                Optimizer           & Adam                 & Adam                 & Adam                 & Adam                  \\
                Betas               & (0.9,0.999)          & (0.9,0.999)          & (0.9,0.999)          & (0.9,0.999)           \\
                Scheduler           & Cosine Annealing     & Cosine Annealing     & Cosine Annealing     & Cosine Annealing      \\
                \bottomrule
            \end{tabularx}
        }
    \end{threeparttable}
    \label{tab:ap_hyperparameter}
\end{table*}
\subsection{Training Details}
In this paper, all experiments are based on the PyTorch framework and running on Nvidia RTX A100 or Nvidia RTX 4090 GPUs. The N-Caltech101 dataset is loaded via SpikingJelly\footnote{https://github.com/fangwei123456/spikingjelly}. Table \ref{tab:ap_hyperparameter} gives the hyperparameter settings for the experiments. In TRT, $\lambda$ controls the initial strength of the regular term and $\delta$ controls the decay rate, the trend of regularization intensity over time is jointly influenced by these two hyperparameters. We have studied the effect of different $\delta$ values in the ablation study section. Furthermore, we find that combining a high $\lambda$ and a low $\delta$ can harm the mid-term training stages. As a result, we suggest using 0.25 as the base value for $\delta$, which should be adjusted based on the magnitude of $\lambda$.

\section{Supplemental Results}
\label{sec:ap_supplemental_results}
\begin{figure}[!ht]
    \centering
    \begin{subfigure}[t]{0.3\columnwidth}
        \centering
        \includegraphics[width=\linewidth]{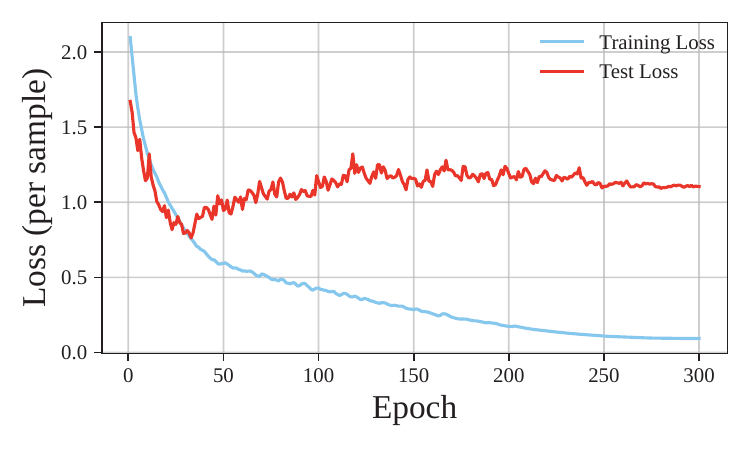}
        \captionsetup{justification=centering}
        \caption{TRT ($\eta=0$).}
        \label{fig:ap_learning_curves_trt_eta0}
    \end{subfigure}
    \begin{subfigure}[t]{0.3\columnwidth}
        \centering
        \includegraphics[width=\linewidth]{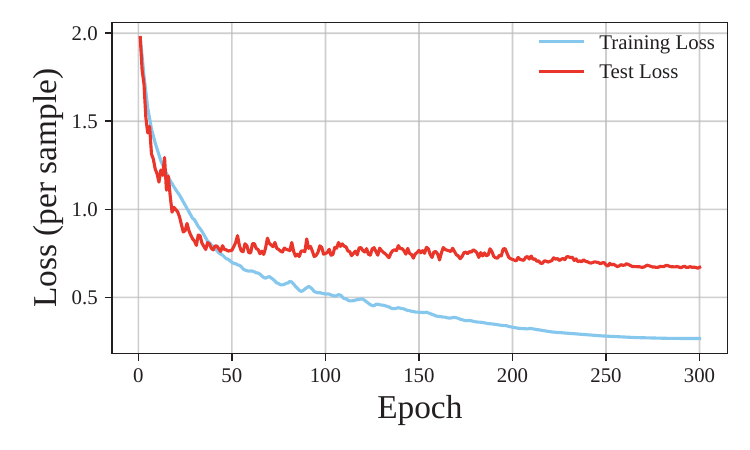}
        \captionsetup{justification=centering}
        \caption{TRT ($\eta=0.1$).}
        \label{fig:ap_learning_curves_trt_eta0.1}
    \end{subfigure}
    \begin{subfigure}[t]{0.3\columnwidth}
        \centering
        \includegraphics[width=\linewidth]{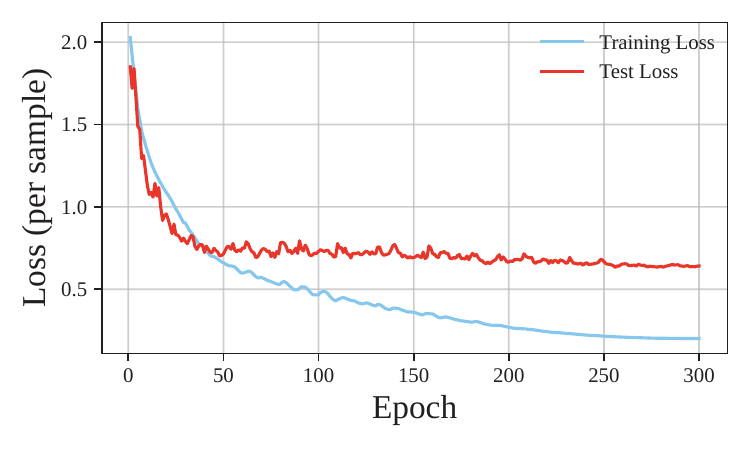}
        \captionsetup{justification=centering}
        \caption{TRT ($\eta=0.05$).}
        \label{fig:ap_learning_curves_trt_eta0.05}
    \end{subfigure}
    \begin{subfigure}[t]{0.3\columnwidth}
        \centering
        \includegraphics[width=\linewidth]{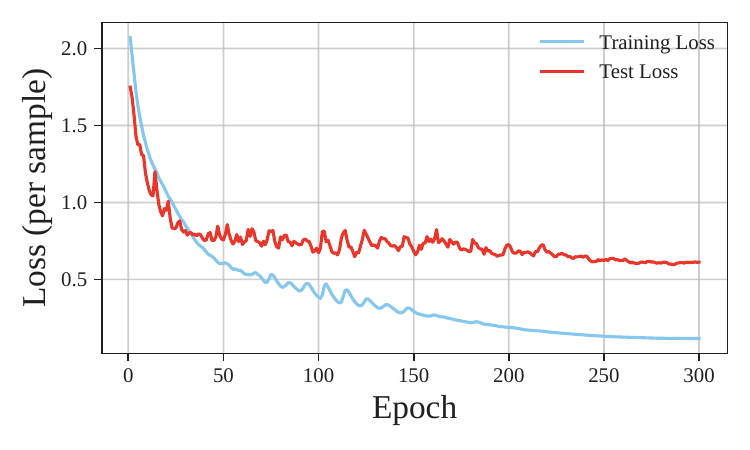}
        \captionsetup{justification=centering}
        \caption{TRT ($\eta=0.01$).}
        \label{fig:ap_learning_curves_trt_eta0.01}
    \end{subfigure}
    \begin{subfigure}[t]{0.3\columnwidth}
        \centering
        \includegraphics[width=\linewidth]{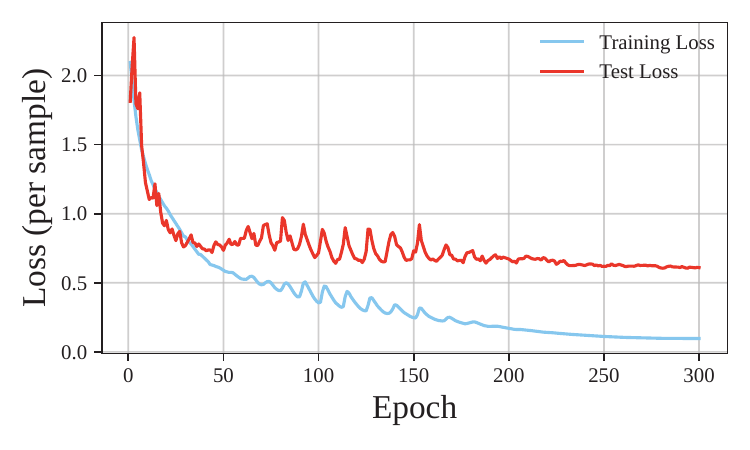}
        \captionsetup{justification=centering}
        \caption{TRT ($\eta=0.005$).}
        \label{fig:ap_learning_curves_trt_eta0.005}
    \end{subfigure}
    \begin{subfigure}[t]{0.3\columnwidth}
        \centering
        \includegraphics[width=\linewidth]{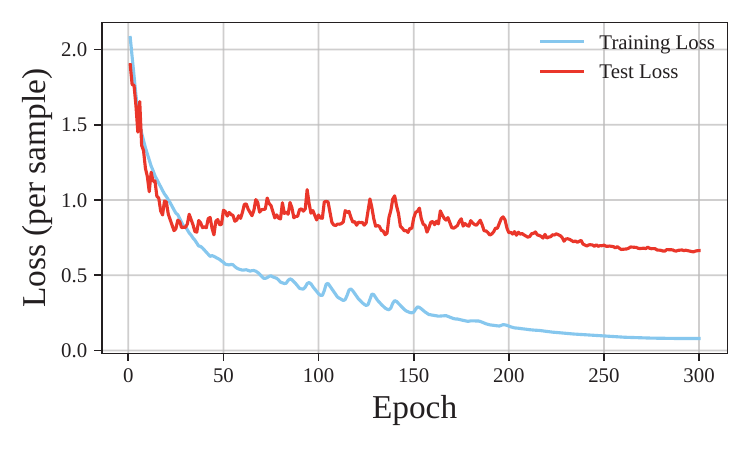}
        \captionsetup{justification=centering}
        \caption{TRT ($\eta=0.001$).}
        \label{fig:ap_learning_curves_trt_eta0.001}
    \end{subfigure}
    \begin{subfigure}[t]{0.3\columnwidth}
        \centering
        \includegraphics[width=\linewidth]{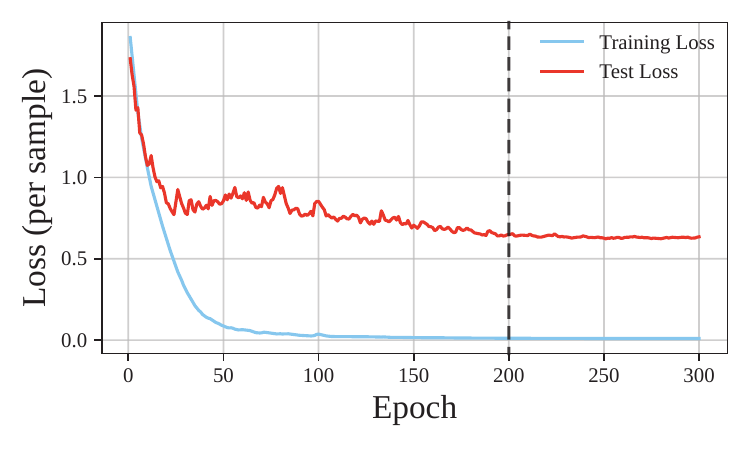}
        \caption{TET.}
        \label{fig:ap_learning_curves_tet}
    \end{subfigure}
    \begin{subfigure}[t]{0.3\columnwidth}
        \centering
        \includegraphics[width=\linewidth]{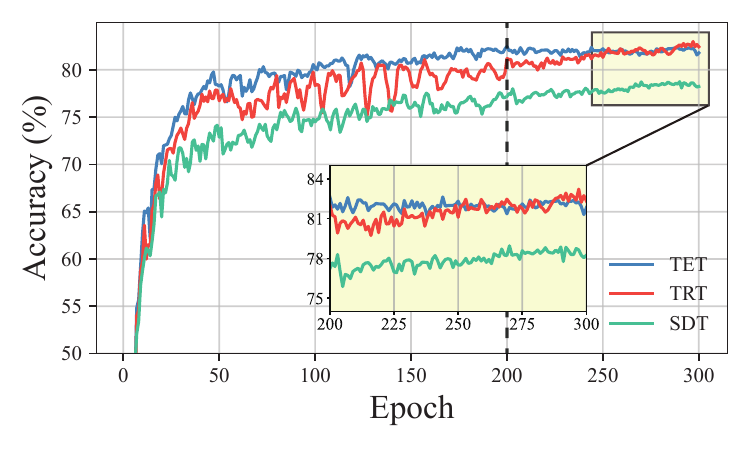}
        \caption{Accuracy curves.}
        \label{fig:ap_acc_curves}
    \end{subfigure}
    \captionsetup{font=normalsize,justification=justified,singlelinecheck=false,width=\columnwidth}
    \caption{Learning curves and accuracy curves on DVS-CIFAR10 dataset. Dash lines in (e) and (f) denote the 200th epoch.}
    \label{fig:ap_dvscifar10_learning_curve_trt}
\end{figure}
\subsection{Learning Curves on Neuromorphic Datasets}
In this part, we visualize the learning curves and the accuracy curves on DVS-CIFAR10 dataset \citep{li2017dvscifar10} (Figure. \ref{fig:ap_dvscifar10_learning_curve_trt}).
Figure. \ref{fig:ap_learning_curves_trt_eta0}-\ref{fig:ap_learning_curves_trt_eta0.001} shows the learning curves of TRT under different values of $\eta$, which controls the ratio of the term $\mathcal{L}_\mathrm{MSE}$. When $\eta$ is set to $1e-3$ (Figure. \ref{fig:ap_learning_curves_trt_eta0.001}), which is the best setting for TRT, the learning curves exhibit a continuous decline in training loss approaching zero, while the test loss stabilizes after an initial decrease, this indicates that TRT can effectively suppress overfitting. In contrast, TET \citep{deng2022tet} shows a sharp initial drop in training loss in the early stages and then fluctuates at lower values, while the testing loss decreases slowly and remains between $0.5-0.7$ after the 200th epoch (Figure. \ref{fig:ap_learning_curves_tet}). The accuracy curves exhibit the same characteristic pattern (Figure. \ref{fig:ap_acc_curves}). This divergence indicates that TET drives the model to fit the training data quickly in the early stages of training, while the test loss fails to decrease correspondingly, demonstrating evident overfitting behavior. This observation reveals that TET exhibits limited capability in mitigating overfitting compared to TRT.
\section*{Code Availability}
The code of this paper is available at \\ \url{https://github.com/ZBX05/Temporal-Regularization-Training}.

\section*{Acknowledgment}
This work is supported by the National Natural Science Foundation of China (Grant No. 12401627). The authors would like to thank the Ph.D candidate Zimeng Zhu from Beihang University for his constructive suggestions on reviewing this manuscript.

\bibliographystyle{elsarticle-harv}
\bibliography{references}






\end{document}